\documentclass{Interspeech}

\usepackage{hyperref}
\usepackage{multirow}
\usepackage{color, colortbl}
\definecolor{Gray}{gray}{0.95}

\interspeechcameraready

\title{An Effective Training Framework for Light-Weight Automatic Speech Recognition Models}

\author[affiliation={1}]{Abdul}{Hannan}
\author[affiliation={2}]{Alessio}{Brutti}
\author[affiliation={3}]{Shah}{Nawaz}
\author[affiliation={4}]{Mubashir}{Noman}

\affiliation{}{University of Trento}{Italy}
\affiliation{}{Fondazione Bruno Kessler}{Italy}
\affiliation{}{Johannes Kepler University}{Austria}
\affiliation{}{Mohamed bin Zayed University of Artificial Intelligence}{U.A.E.}
\email{}

\usepackage{comment}

\begin{document}

\maketitle

\begin{abstract}
Recent advancement in deep learning encouraged developing large automatic speech recognition (ASR) models that achieve promising results while ignoring computational and memory constraints. However, deploying such models on low resource devices is impractical despite of their favorable performance. Existing approaches (pruning, distillation, layer skip etc.) transform the large models into smaller ones at the cost of significant performance degradation or require prolonged training of smaller models for better performance. To address these issues, we introduce an efficacious two-step representation learning based approach capable of producing several small sized models from a single large model ensuring considerably better performance in limited number of epochs. 
Comprehensive experimentation on ASR benchmarks reveals the efficacy of our approach\footnote[1]{Our code is available at: \textcolor{blue}{https://github.com/hannabdul/etf4asr}}, achieving three-fold training speed-up and up to $12.54\%$ word error rate improvement.


\end{abstract}

\vspace{2mm}
\section{Introduction}
Accelerated progress in Internet of Things (IoT) \cite{farahani2023towards} and edge-cloud continuum \cite{liang2023wh} demands significant reduction in the neural architecture sizes for smooth operation with minimal computational and memory burden. Several efforts have been directed to create user and device-personalized models with adaptable model sizes while minimizing the latency using different techniques. Contemporary works aims at sparsification of deep models using pruning \cite{10.1145/3534678.3539260, wei2022model} that primarily locate the intermediate representation (or input) that if removed, would minimally affect the model's output. 
Others either employ low-precision data and models \cite{nguyen20c_interspeech} or opt for low-rank approximation methods \cite{winata2020lightweight, lirias3769084}, or distill the knowledge from a large teacher model to a student model using a distribution's divergence-based loss such as Kullback–Leibler divergence, Jensen-shannon divergence, Wasserstein distance, etc \cite{DBLP:journals/corr/HintonVD15, watanabe2017student, fukuda2017efficient, takashima2019investigation, kurata2020knowledge, yoon2021tutornet, zhao2022knowledge, yoon2024cons}. However, applying pruning solely leads to substantial performance degradation, directing the researchers to use pruning in conjunction with other approaches \cite{wei2022model, zhen2021sparsification, xiao24b_interspeech} to reduce the affect on performance up to some extent.
Alternatively, few works utilized conditional computing approaches, such as Early Exit \cite{DBLP:journals/corr/abs-2309-09546}, Layer dropping \cite{hannan2024, zaiem2023fine} etc., to reduce the computational load at inference time either by exiting at early stage or skipping some parts of the network. This dynamic behavior comes at the cost of computational flow alteration and addition of new modules, raising the computational and memory requirements.
In addition, dynamic dropping of layers leads to considerable reduction in ASR performance.

Curating resource-friendly architectures with the ideal performance is the underlying objective of machine learning, and is featuring new paradigms like tiny machine learning (tiny ML), green ML etc. Developing large models with billions of parameters to gain a point over counterparts might not be the right way to improve task-specific performance. Not only these architecture results in tons of carbon dioxide emission and consume a staggering amount of energy until convergence, the level of resource consumption of such large models does not align with these paradigms as well as their deployment on end-user devices is practically unviable. 
Similarly, compressing large models being undoubtedly effective, suffers from the drawback of inefficient resource utilization in addition to noticeable performance degradation. 
Moreover, training small and device-conditioned models every time for substantial number of epochs is also unfeasible. The reason is that small models demands prolong training time and large number of epochs to converge with subpar performance level. 
Therefore, it is desired to build smaller architectures capable of achieving favorable performance as well as requiring reduced computational and memory load. 

In this regard, we propose an effective approach that exploits the knowledge of large reference model to obtain tiny models capable of performing favorably while requiring low computational resources. To summarize, our contributions are as below.

\begin{itemize}
    \item We propose a two-step based approach to train small ASR models that achieve promising results on ASR task.
    \item The introduced approach enables us to train several small and tiny models requiring less compute resources while providing promising performance compared to equivalent small models.
    \item We perform extensive experiments on two challenging ASR benchmarks demonstrating the effectiveness of proposed approach.
\end{itemize}



\begin{figure*}[!t]
    \centering
    \includegraphics[width=0.9\linewidth]{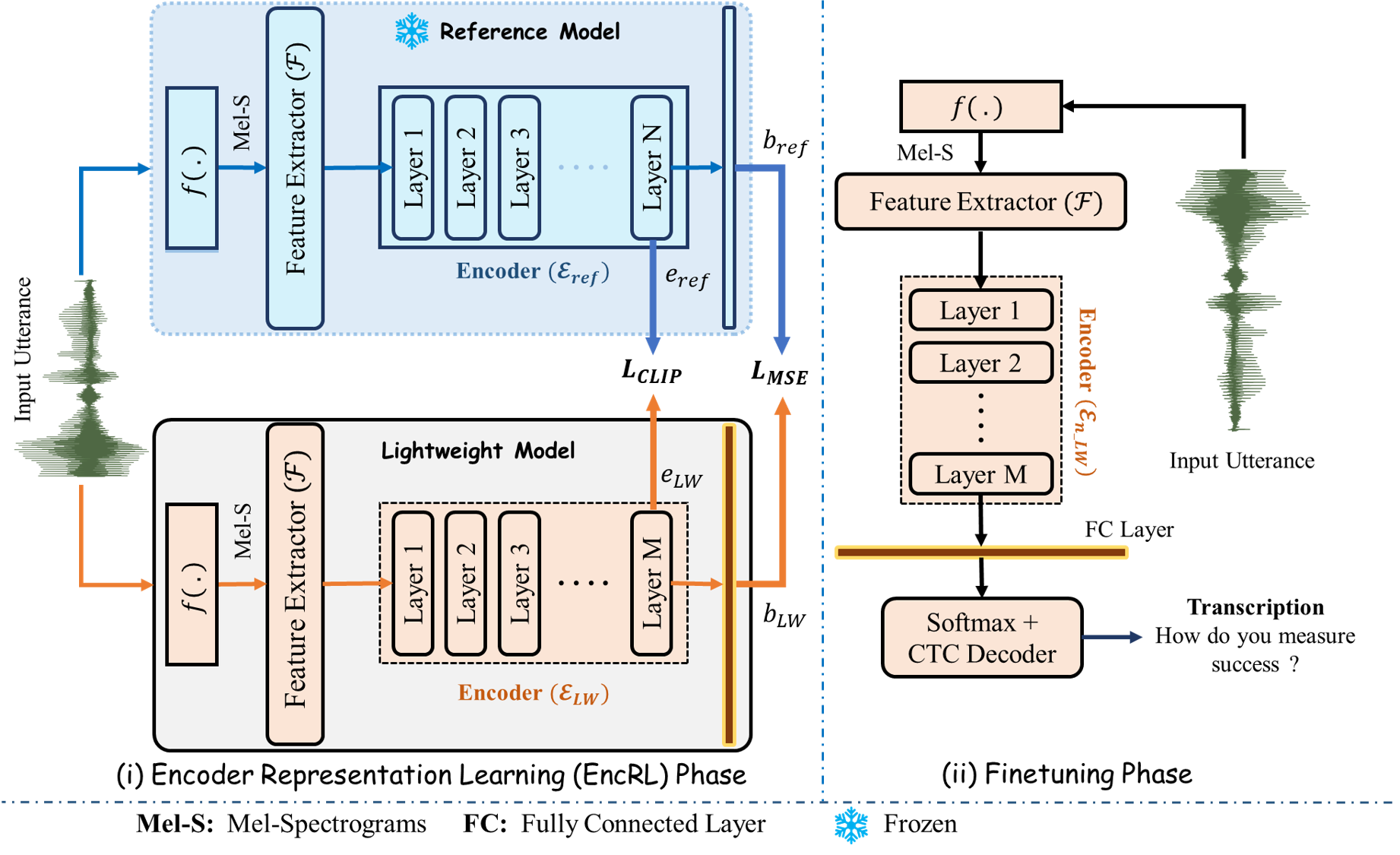}
    \caption{Here we illustrate the overall framework of the proposed method. (i) In the feature learning phase, the input utterance is fed to a large ASR model (reference model) to extract the features $e_{ref}$ and $b_{ref}$. Afterwards, the same input is passed through the smaller model to obtain features $e_{LW}$ and $b_{LW}$. To learn the knowledge of reference model, MSE loss is used between the outputs ($b_{ref}$ and $b_{LW}$) of the classifiers of the reference and the small model. Symmetric cross-entropy loss is used on the features $e_{ref}^{N}$ and $e_{LW}^{M}$ to align the feature spaces. 
    (ii) After transferring the knowledge to the encoder of light-weight model, CTC decoder is integrated and the model is finetuned to transcribe the input speech.}
    \label{fig:framework}
\end{figure*}
\vspace{-1mm}
\section{Background}
\noindent \textbf{Representation Learning using Mask Auto-Encoders: } Mask auto-encoders \cite{He_2022_MAE, noman2024rethinking} are self-supervised frameworks that are utilized to learn rich semantic features from large amount of unlabeled data. Mask auto-encoder masks the large portion of its input $X \in \mathbb{R}^{H \times W}$, and feeds the unmasked patches to the encoder to obtain encoded features $X_e$. 
In decoder, learnable features are appended to the encoded features $X_e$ which are forwarded to the decoder layers to obtain the reconstructed feature representations $X_r$. Mask auto-encoder utilizes the mean squared error loss (MSE) to measure the reconstruction performance and enable the model to learn meaningful feature representations in self-supervised manner.

\noindent \textbf{Knowledge Distillation: } Knowledge distillation is a technique utilized to transfer the knowledge of a large or teacher model to a student model. Typically, the teacher model is trained on large set of available data and the knowledge is transferred to the student model by means of distillation loss. Existing works \cite{hinton2015distillingknowledgeneuralnetwork, knowledge_distill_acoustic} have utilized the cross-entropy loss having soft target distributions to distill the knowledge given as:

\begin{equation}
    c_i = \frac{e^{z_i/T}}{\Sigma_j e^{z_j/T }}
\end{equation}

where $c_i$ is the class probability of $ith$ class, $z_i$ is the logit and $T$ is the temperature to control the distillation. 
This teacher-student architecture is trained in an end-to-end fashion. Given that we have several student networks, it is desired to train each student model using the knowledge distillation framework enabling the student model to perform well on the specific task. 

In addition to the temperature controlled cross-entropy loss, MSE loss is also utilized to distill knowledge for classification task \cite{ijcai2021p362}. However, MSE is typically used in mask auto-encoder frameworks to learn meaningful representations from huge amount of unlabelled data. 
Inspired from these, we propose a two-step approach that first learns the semantically rich features from the encoder of large reference model followed by finetuning tiny models for limited number of epochs.


\section{Proposed Framework} \label{prop_method}
To mitigate the performance limitations and facilitate fast training, we propose a two-step feature learning based framework in Figure \ref{fig:framework}. The proposed framework requires: (i) \textit{One time feature representation learning} to mimic the enriched representation of the large Reference model , (ii) Finetuning for limited number of epochs, outperforming the similar sized models trained from scratch for greater number of epochs.

Given a dataset $\mathcal{D} = \{ (a_d^i, t_d^i) \}_{i=1} ^ {K}$ where $a_d^i$ and $t_d^i$ are the corresponding audio-transcription pair of the $i^{th}$ instance with $K$ total instances, we train a reference model $\mathcal{M}_{ref}$ consisting of $N$ encoder blocks for $Z$ number of epochs. 
This trained model provides the \textit{baseline results} as well as serves as the reference model for representation learning phase.

\subsection{Encoder's Representation Learning (EncRL)} 
\label{method_pretraining}
In this phase, our objective is to enable the light-weight model $\mathcal{M}_{LW}$ to learn the information from the deep reference model $\mathcal{M}_{ref}$ that $\mathcal{M}_{ref}$ has learned by training on a large amount of data for a greater number of epochs $Z$. To this end, reference model $\mathcal{M}_{ref}$ and light-weight model $\mathcal{M}_{LW}$ are created having $N$ and $M$ number of layers, respectively, where $M \leq \frac{N}{2}$. 
Subsequently, we take the output features $e_{LW}$ and $e_{ref}$ of the last layer of the corresponding encoders and utilize the symmetric cross-entropy $\mathcal{L}_{CLIP}$ loss to align the feature representations as shown in Figure~\ref{fig:framework}. The $\mathcal{L}_{CLIP}$ loss minimizes the distance between the feature representations ($e_{LW}$ and $e_{ref}$) of the same samples while maximizing for the different samples. 
Similarly, to enforce the light-weight model $\mathcal{M}_{LW}$ to produce the similar output as the reference model $\mathcal{M}_{ref}$ generates, we utilize mean squared error ($\mathcal{L}_{MSE}$) loss on the feature embeddings $b_{LW}$ and $b_{ref}$ of the two models. This loss combination will encourage the light-weight model $\mathcal{M}_{LW}$ to learn rich semantic information from the model $\mathcal{M}_{ref}$. The total loss is given as: 



\begin{equation}
    \mathcal{L}_{EncRL} = \mathcal{L}_{CLIP} + \mathcal{L}_{MSE}
\end{equation} \vspace{-4.5mm}
\begin{equation} 
    \mathcal{L}_{CLIP} = \max_{i \neq j} \Bigl( \min_{i = j} \Bigl( \sum_{i, j}^{B} e_{ref}^{i} \odot e_{LW}^{j} \Bigl) \Bigl)
\end{equation} \vspace{-4.5mm}
\begin{equation} 
    \mathcal{L}_{MSE} = \frac{1}{B}\sum_{i}^{B} ( b_{ref}^{i} - b_{LW}^{i} )^2
\end{equation}

where $\odot$ represents element-wise multiplication. 

During this phase, the reference model $\mathcal{M}_{ref}$ is kept frozen and the transcription decoder is not utilized. 
We perform one-time feature representation learning of the light-weight model $\mathcal{M}_{LW}$ by training for  $\frac{2\times Z}{3}$ epochs, enabling it to learn the enriched representation of $\mathcal{M}_{ref}$. 
Afterwards, we finetune the model $\mathcal{M}_{LW}$ for a few epochs as explained in the next section. 


\begin{table}[t]
\centering
\caption{WER evaluated on test-clean and test-other splits of LibriSpeech dataset. First row indicates the large ASR model trained for \textit{n} number of epochs. * represents that the smaller model is trained for \textit{n} number of epochs similar to large model. $\dagger$ represents the results of a Conformer model from \cite{DBLP:journals/corr/abs-2309-09546} with twice number of attention heads and doubled feed-forward dimension in conformer submodules. $\times$ represents not reported.}
\label{tab:res-libri}
\setlength{\tabcolsep}{2.5pt}
\begin{tabular}{lccccc}
\toprule
\textbf{Model} & \textbf{Epochs} & \textbf{\begin{tabular}[c]{@{}c@{}c@{}}\# of \\Encoder\\Layers\end{tabular}} & \textbf{\begin{tabular}[c]{@{}c@{}}Params \\(M)\end{tabular}} & \multicolumn{2}{c}{\textbf{WER (\%)}} \\ \cmidrule{5-6}
& & & & \textbf{clean} & \textbf{other} \\ \midrule

Conformer (ref) & 150 & \multirow{2}{*}{12} & 18.60 & 5.60 & 14.38 \\
Conformer$^{\dagger}$ & $\times$ &  & 31.601 & 6.5 & 17.7\\

\midrule
Conformer$^{*}$ & 150 & \multirow{3}{*}{6} & 9.462 & 8.61 & 20.97\\
Conformer$^{\dagger}$ & $\times$ &  & 15.963 & 7.6 & 20.0\\

Conformer (ours) & 50 &  & 9.462 & \textbf{6.27} & \textbf{17.47}\\

\midrule
Conformer$^{*}$ & 150 & \multirow{3}{*}{4} & 6.416 & 8.91 & 22.18\\
Conformer$^{\dagger}$ & $\times$ &  & 10.750 & 9.8 & 24.3\\

Conformer (ours) & 50 &  & 6.416 & \textbf{7.95} & \textbf{21.82}\\

\midrule
Conformer$^{*}$ & 150 & \multirow{3}{*}{2} & 3.370 & 15.57 & \textbf{32.57} \\
Conformer$^{\dagger}$ & $\times$ &  & 5.537 & 17.6 & 36.1\\

Conformer (ours) & 50 &  & 3.370 & \textbf{14.68} & 33.70\\

\bottomrule 
\end{tabular}
\end{table}

\subsection{Finetuning phase} \label{method_finetuning}
After the feature representation learning phase, we integrate the Connectionist Temporal Classification (CTC) decoder in the light-weight model $\mathcal{M}_{LW}$. Then, we initialize the model $\mathcal{M}_{LW}$ with the weights of the representation learning phase. 
We feed the input audio utterances to the model $\mathcal{M}_{LW}$ and generate the output transcriptions. We finetune the model for $Z/3$ number of epochs. During finetuning, we utilize the connectionist temporal classification loss to measure the performance of the model. 
Similar to the model $\mathcal{M}_{LW}$, we further finetune the tiny models $\mathcal{M}_{n\_LW}$ having number of layers less than $M$. Specifically, we finetune the model $\mathcal{M}_{n\_LW}$ having encoder layers $n$ where $n \in [2,4,...,M]$. This enables the smaller models to perform favorably without requiring training for a large number of epochs. 

For instance, given a mini-batch $B$ in the dataset $\mathcal{D}$, the model processes the audio inputs $a^i$ producing the corresponding embedding $b^i$, followed by a \textit{Softmax} layer that converts the raw logits into the respective token probabilities $p^i$ as illustrated in Figure~\ref{fig:framework}. The tokens are forwarded to a built-in CTC decoder transforming the incoming data into the final transcription $t^i_{pred}$. Finally, the model's transcription is optimized by quantification of CTC loss by comparing against the ground-truth transcription.
\begin{equation}
    \begin{split}
        t^i_{pred} = CTC\_Decoder \hspace{1mm} \big(Softmax(b^i)\big) \\ 
        \mathcal{L}_{ctc} = f_{CTC}(t^i - t^i_{pred}) \hspace{13mm}
    \end{split}
\end{equation}


\section{Experiments and Results}
\noindent \textbf{Implementation Details:} The model is trained on NVIDIA $A40$ GPU with a batch size of 64 using AdamW optimizer with $L2$ normalization of $1e^{-6}$ and $10000$ warmup iterations followed by exponentially decreasing the learning rate till the end of training. 
For each audio input, mel-spectrograms (using $80$ mel-filterbanks) are computed for $20ms$ window length with a hop-length of $10ms$ along with SentencePiece byte-pair-encoding (BPE-256) tokenizer \cite{kudo-richardson-2018-sentencepiece}. To suppress the over-fitting issue, we applied \textit{SpecAug} augmentation \cite{park19e_interspeech} with frequency and time masking of $27$ and $80$ samples.

\noindent \textbf{Architecture:} We adapted open-sourced Conformer (small) architecture with feed-forward dimension of $256$ and $12$ encoder layers for the reference model. Consequently, the number of encoder layers $M$ for light-weight model are $6$. Furthermore, we replaced the 3-layered LSTM based decoder with a linear projection layer, whose output is forwarded to the CTC decoder for transcription.

\noindent \textbf{Datasets:} We evaluated our method on two challenging and publicly available ASR datasets: \textbf{LibriSpeech} \cite{panayotov2015librispeech} comprising of $\approx$$1000$ hours of read-aloud audiobooks and \textbf{TED-LIUM} (release-3) \cite{hernandez2018ted} consisting of $\approx$$452$ hours of transcribed English TED talks. We employed Word Error Rate (WER) metric to quantify the transcription errors.

\subsection{Results}
\begin{table}[t]
\centering
\caption{WER evaluated on TEDLIUM-v3 dataset (test split)}
\label{tab:res-ted}
\setlength{\tabcolsep}{4.5pt}
\begin{tabular}{lcccc}
\toprule
\textbf{Model} & \textbf{\begin{tabular}[c]{@{}c@{}}\# of Encoder\\Layers\end{tabular}} & \textbf{\begin{tabular}[c]{@{}c@{}}Params \\(M)\end{tabular}} & \textbf{WER (\%)} \\ \midrule

Conformer (ref) & \multirow{2}{*}{12} & 18.60 & 15.28 \\
Conformer$^{\dagger}$ &  & 31.601 & 16.4\\

\midrule
Conformer$^{*}$ & \multirow{3}{*}{6} & 9.462 & 21.82\\
Conformer$^{\dagger}$ &  & 15.963 & 25.5\\

Conformer (ours) &  & 9.462 & \textbf{19.72}\\

\midrule
Conformer$^{*}$ & \multirow{3}{*}{4} & 6.416 & 49.30\\
Conformer$^{\dagger}$ &  & 10.750 & 35.4\\

Conformer (ours) &  & 6.416 & \textbf{22.86}\\

\bottomrule 
\end{tabular}
\end{table}

\noindent \textbf{Comparison with small models:} We compare the proposed framework with existing light-weight models, as well as smaller models that are trained for $Z$ number of epochs. Table \ref{tab:res-libri} reveals that our approach achieves superior performance as compared to the other smaller models for various model sizes (different number of encoder layers). For instance, our method obtains WER of $6.27\%$ and $14.68\%$ for the model having $6$ and $2$ encoder layers on LibriSpeech test-clean split, whereas the smaller models that are trained for greater epochs achieve WER of $8.61\%$ and $15.67\%$ respectively. Similarly, our approach achieves reasonable improvement in WER score of $3.5\%$ and $0.36\%$ for model having $6$ and $4$ encoder layers on LibriSpeech test-other split. Likewise, on Tedlium (v3) test split, our approach obtains absolute improvement of $2.1\%$ and $26.44\%$ in WER metric as compared to model trained from scratch.

In addition, to train $W$ models of various sizes, our method requires $(W \times Z )/3$ training epochs 
compared to the other smaller models that require $W \times Z$ epochs, 
achieving $3 \times$ faster training and better performance.
In short, our approach offers flexibility 
to create several light-weight and small-sized models requiring considerably less training time 
and improved performance.

\noindent \textbf{Comparison with Dynamic models:}
To further validate our framework's efficiency, we compare the light-weight models finetuned with our approach against conditional computing approaches such as Early Exit (EE) \cite{DBLP:journals/corr/abs-2309-09546} and Random Layer Dropping (RD) \cite{hannan2024}. Table \ref{tab:comp_dyn} depicts the performance superiority of our framework, surpassing EE and RD approaches for various model sizes. 
It is worth mentioning that EE and RD enables a single model to dynamically adjust the model's size, hence, suitable for storage-constrained environments while compromising on the final accuracy. In contrast, our framework produces multiple light-weight models suitable for different resource settings with considerably better performance as compared to the counterparts. 
\begin{table}[t]
\centering
\caption{WER (\%) comparison with conditional computation strategies on LibriSpeech test-clean split.}
\label{tab:comp_dyn}
\setlength{\tabcolsep}{6pt}
\begin{tabular}{cccc}
\toprule
\textbf{\textbf{\begin{tabular}[c]{@{}c@{}}\# of Encoder\\Layers\end{tabular}}} & \multicolumn{3}{c}{\textbf{WER (\%)}$_{\textcolor{blue}{Params (M)}}$} \\ \cmidrule{2-4} 
& Ours & \begin{tabular}[c]{@{}c@{}}Early Exit\\ \cite{DBLP:journals/corr/abs-2309-09546} \end{tabular} & \begin{tabular}[c]{@{}c@{}}Random \\Dropping \cite{hannan2024} \end{tabular} \\ \midrule

6 & \textbf{6.27}$_{\hspace{0.75mm} \textcolor{blue}{9.46}}$ & 6.8$_{\hspace{0.75mm} \textcolor{blue}{15.96}}$ & 7.98$_{\hspace{0.75mm} \textcolor{blue}{15.96}}$\\
4 & \textbf{7.95}$_{\hspace{0.75mm} \textcolor{blue}{6.41}}$ & 11.6$_{\hspace{0.75mm} \textcolor{blue}{10.75}}$ & 12.54$_{\hspace{0.75mm} \textcolor{blue}{10.75}}$\\
2 & \textbf{14.68}$_{\hspace{0.75mm} \textcolor{blue}{3.37}}$ & 23.9$_{\hspace{0.75mm} \textcolor{blue}{5.53}}$ & 43.62$_{\hspace{0.75mm} \textcolor{blue}{5.53}}$\\

\bottomrule 
\end{tabular}
\vspace{-3mm}
\end{table}

\noindent \textbf{Why EncRL is required?} 
We performed the finetuning of smaller model having 6 encoder layers while initializing the weights from the final states of $\mathcal{M}_{ref}$. We observe that despite of intializing the weights from $\mathcal{M}_{ref}$ model, the light-weight model struggles to provide better performance and achieves a WER of $7.41\%$. Whereas our approach provides better performance by achieving WER of $6.27\%$. 

\noindent \textbf{Vitality of Finetuning:} To investigate the requirement of finetuning, 
we integrated the CTC decoder to the light-weight model obtained after representation learning. We then evaluated the model on LibriSpeech test-clean split. It was observed that the model failed to recognize the audio signal resulting in extremely high WER of ($\approx 95\%$) highlighting the need of finetuning phase.

\subsection{Ablations} \label{ablations}
\subsubsection{Quantifying the efficiency of Loss functions}
We employed various loss functions during the representation learning phase including CLIP, MAE, MSE, and their combinations. 
Table \ref{tab:ablation-loss} enlists the obtained WER for each loss function, highlighting that combination of CLIP and MSE loss 
provide the rich feature representations 
which enable the model to obtain best performance after finetuning phase achieving WER score of 6.27\%. 

\begin{table}[t]
\centering
\caption{Evaluating the effect of different Loss functions used during Encoder's Representation Learning phase. \# of encoder layers used in these experiments is 6.}
\label{tab:ablation-loss}
\setlength{\tabcolsep}{20pt}
\begin{tabular}{lc}
\toprule
\textbf{Loss Function} & \textbf{WER (\%)} \\ \midrule

CLIP only & 9.06\\
MAE only & 6.40\\
MSE only & 6.36\\
CLIP + MAE & 6.42\\
CLIP + MSE & \textbf{6.27}\\

\bottomrule 
\end{tabular}
\vspace{-4mm}
\end{table}

\subsubsection{Do we need separate EncRL for various sizes of model?} \vspace{-1mm}
To answer this question, we performed \textit{EncRL} for a model having $4$ encoder layers followed by finetuning for $Z/3$ epochs, and compared it with the finetuned model that is initialized with the last 4 layers of encoder's state dictionary of \textit{EncRL} with $6$ layers. We observed slight improvement of $0.53\%$ in WER when we separately perform encoder representation learning for 4 encoder layers
demonstrating that the repeating encoder representation learning step does not provide significant improvement in performance but consumes twice as much computational cost. \vspace{-1mm}

\subsubsection{Pruning} \vspace{-1mm}
We applied pruning technique on the reference model to compare its performance against the proposed framework on LibriSpeech test-clean split. To prune the model, we employed two different schemes: (i) pruning conformer module completely by a given amount, (ii) pruning other conformer sub-modules except the convolutional sub-module. For the \textit{former} case, we observed that pruning only $5\%$ of weights significantly degrades the performance, scoring $13.64\%$ WER. The performance reduces further if we increase the pruning amount by $2.5\%$, and totally collapses at pruning amount of $10\%$ giving a WER of $49.44\%$. For the \textit{latter} case, the model shows WER of $10.51\%$ for $7.5\%$ pruned weights which is $13.16\%$ less than we achieved in the former pruning method. Similarly, the WER substantially increases by $11.69\%$ and $23\%$ for pruning amount of $15\%$ and $22.5\%$ highlighting the limitation of pruning to maintain good performance. However, our framework significantly outperforms the pruning schemes, providing a mere WER of $6.27\%, 7.95\%$ and $14.68\%$ for equivalent pruning amount of $50\%, 66.7\%$, and $83.33\%$, highlighting the efficacy of the framework.



\section{Conclusion}
In this work, we proposed a two-step framework to create ultra light-weight models in a resource efficient manner without significantly compromising on the final performance. 
We demonstrated that a reference model can be employed to train a general light-weight encoder-only model that serves as a starting point for multiple light-weight networks. 
This is accomplished by transferring the knowledge of the reference model to the light-weight model through feature alignment and adaptation.  
Later, the model with desired encoder depth is instantiated and initialized with the final state of the representation learning, and finetuned for a handful of epochs on the target dataset. Extensive experimentation on LibriSpeech and Tedlium datasets reveals the efficiency of the proposed framework.

\bibliographystyle{IEEEtran}
\bibliography{mybib}

\end{document}